\documentclass{article}
\usepackage{spconf,amsmath,graphicx,hyperref}
\usepackage{booktabs}
\usepackage{multirow}

\usepackage{amsfonts}     
\usepackage{nicefrac}     
\usepackage{microtype}     
\usepackage{wrapfig}
\usepackage{graphicx}

\usepackage{svg}
\usepackage{colortbl}
\usepackage{amsmath}
\usepackage{amssymb}
\usepackage{color}
\usepackage{multirow}
\usepackage{indentfirst}
\usepackage{threeparttable}
\usepackage{pifont}
\usepackage{makecell}
\usepackage{wrapfig}
\usepackage{longtable}
\usepackage{tablefootnote}
\usepackage{arydshln}

\usepackage{bbding}       

\usepackage{url}          
\usepackage{xcolor}       
\usepackage{changes}

\title{PictOBI-20k: Unveiling Large Multimodal Models in Visual Decipherment for Pictographic Oracle Bone Characters}

\name{Zijian Chen$^{1,2,\dag}$, Wenjie Hua$^{3,\dag}$, Jinhao Li$^{4}$, Lirong Deng$^{5}$, Fan Du$^{6}$, Tingzhu Chen$^{1,\star}$,  Guangtao Zhai$^{1,2,\star}$}
  
\address{$^1$Shanghai Jiao Tong University, $^2$Shanghai AI Lab, $^3$Wuhan University, $^4$East China Normal Unversity \\ $^5$Macao Polytechnic University, $^6$Southern University of Science and Technology \\ $^{\dag}$Equal contribution  \qquad $^{\star}$Corresponding authors}

\begin{document}

\maketitle

\begin{abstract}
Deciphering oracle bone characters (OBCs), the oldest attested form of written Chinese, has remained the ultimate, unwavering goal of scholars, offering an irreplaceable key to understanding humanity’s early modes of production. Current decipherment methodologies of OBC are primarily constrained by the sporadic nature of archaeological excavations and the limited corpus of inscriptions. With the powerful visual perception capability of large multimodal models (LMMs), the potential of using LMMs for visually deciphering OBCs has increased.
In this paper, we introduce PictOBI-20k, a dataset designed to evaluate LMMs on the visual decipherment tasks of pictographic OBCs. It includes 20k meticulously collected OBC and real object images, forming over 15k multi-choice questions. We also conduct subjective annotations to investigate the consistency of the reference point between humans and LMMs in visual reasoning. 
Experiments indicate that general LMMs possess preliminary visual decipherment skills, and LMMs are not effectively using visual information, while most of the time they are limited by language priors. We hope that our dataset can facilitate the evaluation and optimization of visual attention in future OBC-oriented LMMs. The code and dataset will be available at \url{https://github.com/OBI-Future/PictOBI-20k}.
\end{abstract}

\begin{figure}[t]
\begin{center}
\includegraphics[width=1\linewidth]{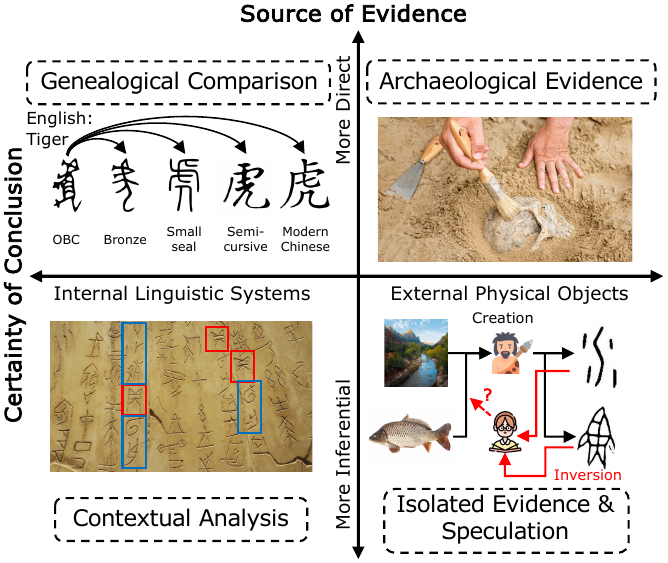}
\caption{Quadrants for contemporary methodologies in the decipherment of ancient scripts: 1) Archaeological evidence, 2) Genealogical comparison, 3) Contextual analysis, and 4) Object-referential method.}
\label{intro}
\end{center}
\end{figure}

\section{Introduction}
Oracle bone character (OBC), as the oldest form of written Chinese, dating to the Shang Dynasty (1250–1050 BC), reflects humanity’s early modes of production, social structures, and the evolution of civilization \cite{chenobi}. Since its discovery in 1899, its decipherment has been a primary objective for scholars. The existing methodologies for OBC deciphering can be broadly classified into four categories, namely archaeological evidence, genealogical comparison, contextual analysis, and object-referential method, as shown in Fig. \ref{intro}. 

From the perspectives of the source of evidence and the certainty of conclusion, archaeological excavations undoubtedly represent the most direct and persuasive method of deciphering. However, its randomness and uncontrollability hinder wide implementation. Recently, generative models have emerged as a powerful tool with record-breaking performance in many applications, including image synthesis \cite{li2025mitigating}, inpainting \cite{li2025obiformer}, and translation.
Sundial-GAN \cite{chang2022sundial} and OBSD \cite{guan2024deciphering} utilize generative adversarial networks (GAN) and a conditional diffusion-based strategy to perform genealogical comparison and bridge OBC with modern Chinese characters for decipherment, respectively. According to the common components within OBCs, Hu et al. \cite{hu2025component} proposed a component-level segmentation-based OBC decipherment method. However, a significant number of OBCs have lost their genealogical lineage, making it impossible to establish a direct correspondence to modern characters, which severely constrains the effective application of these methods.

As large multimodal models (LMMs) have demonstrated their powerful cross-modal information processing capabilities in various fields, researchers are gradually incorporating this technology into the OBC processing tasks \cite{chenobi}. For the deciphering of OBCs, due to the lack of sufficient corpora of OBCs, relying solely on language models for contextual analysis is presently not a tenable proposition. Therefore, the visual-linguistic collaborative understanding-based method becomes the most preferred strategy for deciphering OBCs currently. Specifically, OracleSage \cite{jiang2024oraclesage} introduces hierarchical visual-semantic fusion through progressive fine-tuning of LLaVA’s \cite{liu2023visual} visual backbone and graph-based reasoning mechanisms for interpreting OBC. V-Oracle \cite{qiao2025v} applies principles of pictographic character formation and establishes a multi-stage oracle alignment tuning for improving the basic understanding of OBCs.
However, despite the effective transformation of LMMs that enables them to produce decent descriptions for OBCs, it is difficult to determine textual ground-truth or a perfect description for OBCs, which normally requires the consensus of experts, making it challenging to quantitatively evaluate the model's performance. Furthermore, although a large amount of existing work reports poor decipherment performance for general-purpose LMMs such as GPT-4o \cite{GPT-4o} and Qwen2.5-VL \cite{bai2025qwen25} series, the underlying cause remains unexplored.

\begin{figure*}[t]
\begin{center}
\includegraphics[width=1\linewidth]{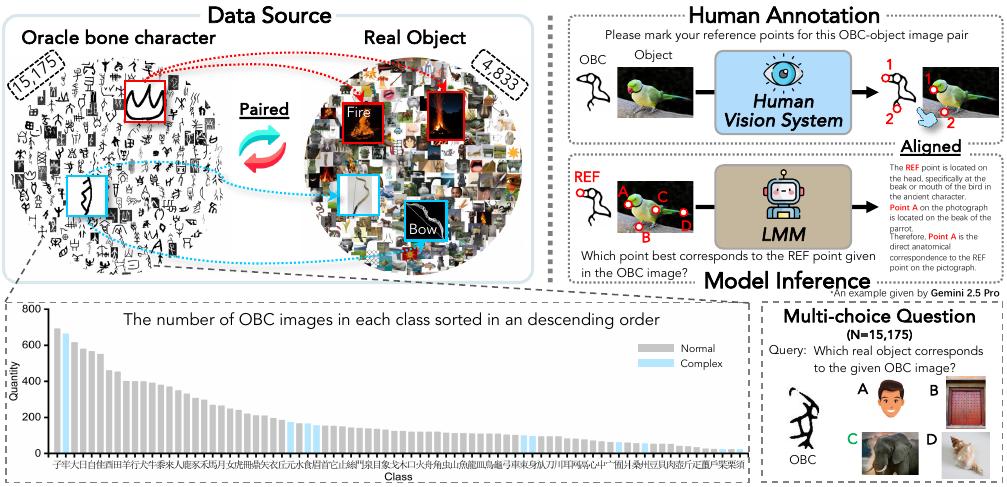}
\caption{Overview of PictOBI-20k. We collect OBC and real object images from 12 sources, covering multiple font appearances and categories. Based on these, we construct 15,175 multi-choice questions for LMM evaluation. Meanwhile, we conduct human annotations for obtaining reference points on OBC-object image pairs.}
\label{overview}
\end{center}
\end{figure*}

To address these problems, we introduce PictOBI-20k, a large-scale OBC-object image dataset dedicated to evaluating the visual decipherment performance of LMMs.
Compared to pure textual decipherment, the visual deciphering of OBC, namely searching for objects that are visually aligned with the OBC, is more consistent with the reverse-engineered version of early character creation methods and is more intuitive in result. 
PictOBI-20k contains 80 types of pictographic OBCs with a total of 15,175 images covering from rubbing to handprinted forms, as well as 4,833 corresponding object images. Based on these, we construct 15k multi-choice questions with ground-truth visual objects (Fig. \ref{overview}). To investigate how humans decipher the OBCs visually, we conduct subjective annotations to obtain reference point maps, which enable the visual alignment consistency experiments between humans and LMMs. 
Experiments show that existing cutting-edge LMMs have not yet reached half the level of humans in aligning pictographic OBCs with real objects.
Moreover, we compare LMMs to three commonly used standalone visual encoders (CLIP-L/14 \cite{radford2021learning}, DINOv2-L/14 \cite{oquab2024dinov2}, and InternViT-300M \cite{zhu2025internvl3}) and diagnose the performance discrepancies.
We anticipate that our work can lay the foundation for the future development and evaluation of vision-centric LMMs for OBC decipherment.

To summarize, the main contributions of this work are:
\begin{itemize}
    \item We construct PictOBI-20k, a large pictograph-specific OBC-object paired image dataset for benchmarking the visual decipherment ability of LMMs.
    \item We conduct subjective annotation to obtain the reference point heatmaps for investigating the alignment between LMMs and humans in deciphering OBCs.  
    \item We evaluate 11 LMMs in the visual decipherment task of pictographic OBCs and discuss the performance of vision backbones separately. Experiments reveal a substantial gap in directly deciphering pictographic OBCs with current LMMs and underscore these models’ oversight of visual representations.
\end{itemize}

\begin{table}[t]
    \centering
    \renewcommand\arraystretch{1}
    \caption{Statistics of the source of OBCs and their corresponding real object images in PictOBI-20k.}
    \resizebox{1\linewidth}{!}{\begin{tabular}{l|l|c|c}
    \hline
        \textbf{Data Category} & \textbf{Source} &\bf Quantity&\textbf{Avg. Reso.} \\ \hline
        \multirow{8}{*}{OBC images}&YinQiWenYuan\tablefootnote{\url{https://www.jgwlbq.org.cn/home}} &761&\multirow{8}{*}{512$\times$512}\\
        & XiaoXueTang\tablefootnote{\url{https://xiaoxue.iis.sinica.edu.tw/}} &1731&\\
        &GuoXueDaShi\tablefootnote{\url{https://www.guoxuedashi.net/}}&35&\\
        &Oracle-241 \cite{wang2022unsupervised}&1994&\\
        &Oracle-50K \cite{han2020self}&5635&\\
        &HUST-OBS \cite{wang2024open}&4539&\\
        &OBI125 \cite{yue2022dynamic}&410&\\
        &OBIdataseIJDH \cite{OBI-IJDH}&70&\\
        \hdashline
        \multirow{4}{*}{Object images}&Freepik\tablefootnote{\url{https://www.freepik.com/}}&\multirow{4}{*}{4833}&\multirow{4}{*}{512$\times$610}\\
        &Pexels\tablefootnote{\url{https://www.pexels.com/zh-cn/}}&&\\
        &Pinterest\tablefootnote{\url{https://www.pinterest.com/}}&&\\
        &Bronze Ware Database\tablefootnote{\url{https://bronze.asdc.sinica.edu.tw/}}&&\\
        \hline
    \end{tabular}}
    \label{datasource}
\end{table}

\section{The PictOBI-20k}
\subsection{Data Curation}
To construct a holistic benchmark specifically dedicated to the visual decipherment of pictographic OBCs, we mainly focus on the font appearances, categories, and visual alignment.
Specifically, we follow the taxonomy of pictographic oracle bone characters studied by Xigui Qiu \cite{xigui1988}, an authority in the field of OBI research in China. 
A total of 80 representative pictographic OBC categories are selected, with 70 normal and 10 complex.
The latter are distinguished by being composite images that rely on contextual elements for clarity, rather than representing objects in isolation.
To ensure the comprehensiveness of the data source, we collect OBC images from three OBC-centric ancient script websites, YinQiWenYuan, XiaoXueTang, and GuoXueDaShi, as well as five open-source OBC datasets, including Oracle-241 \cite{wang2022unsupervised}, Oracle-50k \cite{han2020self}, HUST-OBS \cite{wang2024open}, OBI125 \cite{yue2022dynamic}, and OBIdatasetIJDH \cite{OBI-IJDH}, encompassing both rubbing and handprinted forms. 
As for the corresponding real-world object images, we manually download from Freepik, Pexels, and Pinterest under appropriate CC BY licenses.
Consequently, we obtain 15,175 different OBC images and 4,833 object images, as shown in Fig. \ref{overview}.
See Tab. \ref{datasource} for more details.

\begin{figure}[t]
\begin{center}
\includegraphics[width=1\linewidth]{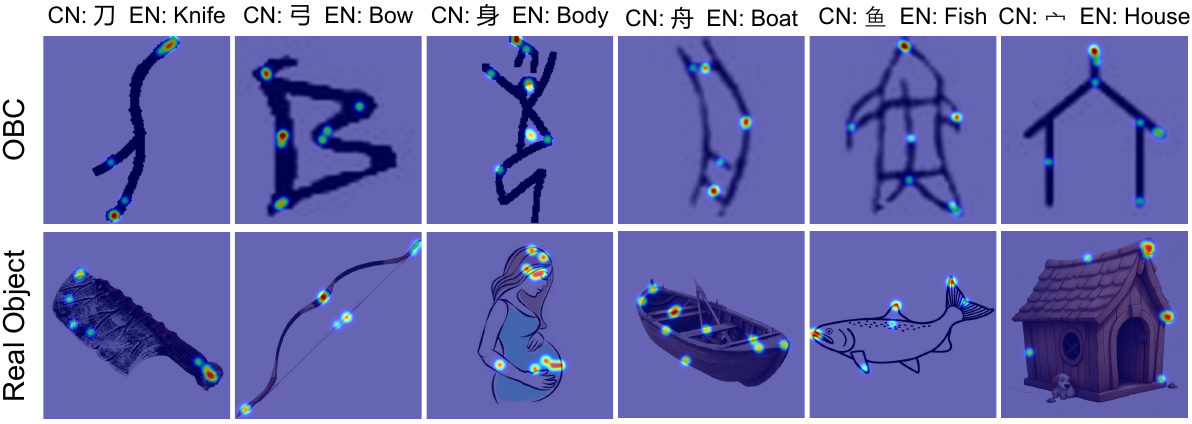}
\caption{Examples of reference point heatmaps.}
\label{annotated}
\end{center}
\end{figure}

\subsection{Inference and Evaluation Strategy}
\noindent
{\bf Multi-Choice Question.}
The main purpose of PictOBI-20k is to measure the performance of general-purpose LMMs in the visual decipherment of OBCs.
Instead of applying a common `Yes-or-No' question that is prone to suffer response preference ({\bf bias}) issues \cite{chen2025just}, we adopt the multi-choice form following many existing practices \cite{qiao2025v,chenobi}. 
As the example shown in Fig. \ref{overview}, the candidate LMM is given a query OBC image and a prompt ``{\it Which real object corresponds to the given OBC image?}", where the interference options are randomly sampled from the other classes.
To mitigate option-position bias, we cyclically rotate the answer options randomly, generating versions with different option orders.

\noindent
{\bf Comparing Visual Alignment with Humans.}  
To explore the inherent visual decipherment potential of LMMs and the visual alignment mechanism of humans, we commence by collecting human feedback on a subset of PictOBI-20k ({\it three pairs for each type}).
Five OBC experts with an ancient philology background were recruited to mark two points on both the OBC image and object image to indicate the corresponding reference regions (Fig. \ref{overview}). Besides, another two irrelevant points are also added to serve as the interference term.
Considering the average resolution of images in PictOBI-20k, the radius of the mark point is set to 8 pixels for good visibility. 
Fig. \ref{annotated} exhibits some examples of reference point heatmaps. We observe a high degree of morphological alignment between pictographic OBCs and real objects, showing the key points of human focusing.
Moreover, we evaluate the visual perception consistency between LMMs and humans using the annotated reference points by prompting ``{\it Which point best corresponds to the REF point given in the OBC image?}"
As a result, 15,175 multi-choice questions and 240 annotated OBC-object image pairs are obtained.

\section{Experiments}
\subsection{Evaluation Setup}

\noindent
{\bf Baselines.} We evaluate a total of 11 LMMs, with a mix of top proprietary models, large and small open-source models, including GPT-4o \cite{GPT-4o}, Gemini 2.5 Pro \cite{gemini25}, Claude 4 Sonnet \cite{claude4}, GLM-4.5V \cite{vteam2025glm45v}, Qwen2.5-VL-\{3B, 7B, 32B, 72B\} \cite{bai2025qwen25}, and InternVL3-\{8B, 38B, 78B\} \cite{zhu2025internvl3}, based on API and their official implementations.
To study the effect of separate visual representations in understanding pictographic OBCs, we explore three common vision backbones, i.e., DINOv2 L/14 \cite{oquab2024dinov2}, CLIP L/14 \cite{radford2021learning}, and InternViT-300M-448px-V2.5 \cite{zhu2025internvl3}.

\noindent
{\bf Implementation Details.} For multiple-choice questions and reference point alignment experiments, we use accuracy and consistency as the metric, respectively.
Drawn from \cite{fu2025hidden}, we compute cosine similarity between the patch features of the reference and the four options for evaluating the vision encoder \cite{fu2025hidden}.
Apart from proprietary LMMs that were deployed via API, the others were run on 4 Nvidia H200 141GB GPUs.

\begin{table}[t]
    \centering
    \renewcommand\arraystretch{1}
    \caption{Accuracy (\%) on PictOBI-20k for the visual decipherment ability of LMMs. The upper part and the lower part are proprietary LMMs and open-source LMMs, respectively. The best and second-best are on {\bf bold} and \underline{underlined}.}
   \resizebox{1\linewidth}{!}{\begin{tabular}{lccc}
    \hline                
        Model&Normal&Complex&Overall\\
        \hline
        {\it Random: 4-choice}&25.00&25.00&25.00\\
        \hdashline
        GPT-4o-2024-11-20&26.31&25.52&26.23\\
        Gemini 2.5 Pro&{\bf 55.22}&\underline{39.44}&{\bf 53.66}\\
        Claude 4 Sonnet-20250514&35.93&25.92&34.94\\
        \hdashline
        GLM-4.5V-106B&33.19&27.11&32.48\\
        Qwen2.5-VL-72B&25.41&24.98&25.36\\
        Qwen2.5-VL-32B&26.77&29.11&32.56\\
        Qwen2.5-VL-7B&26.22&25.72&26.17\\
        Qwen2.5-VL-3B&24.81&24.12&24.76\\
        InternVL3-78B-Instruct&52.29&36.38&50.71\\
        InternVL3-38B-Instruct&\underline{52.71}&{\bf 39.51}&\underline{51.40}\\
        InternVL3-8B-Instruct&33.69&22.25&26.99\\
        \hline
    \end{tabular}}
    \label{performance}
\end{table}

\begin{figure}[t]
\begin{center}
\includegraphics[width=1\linewidth]{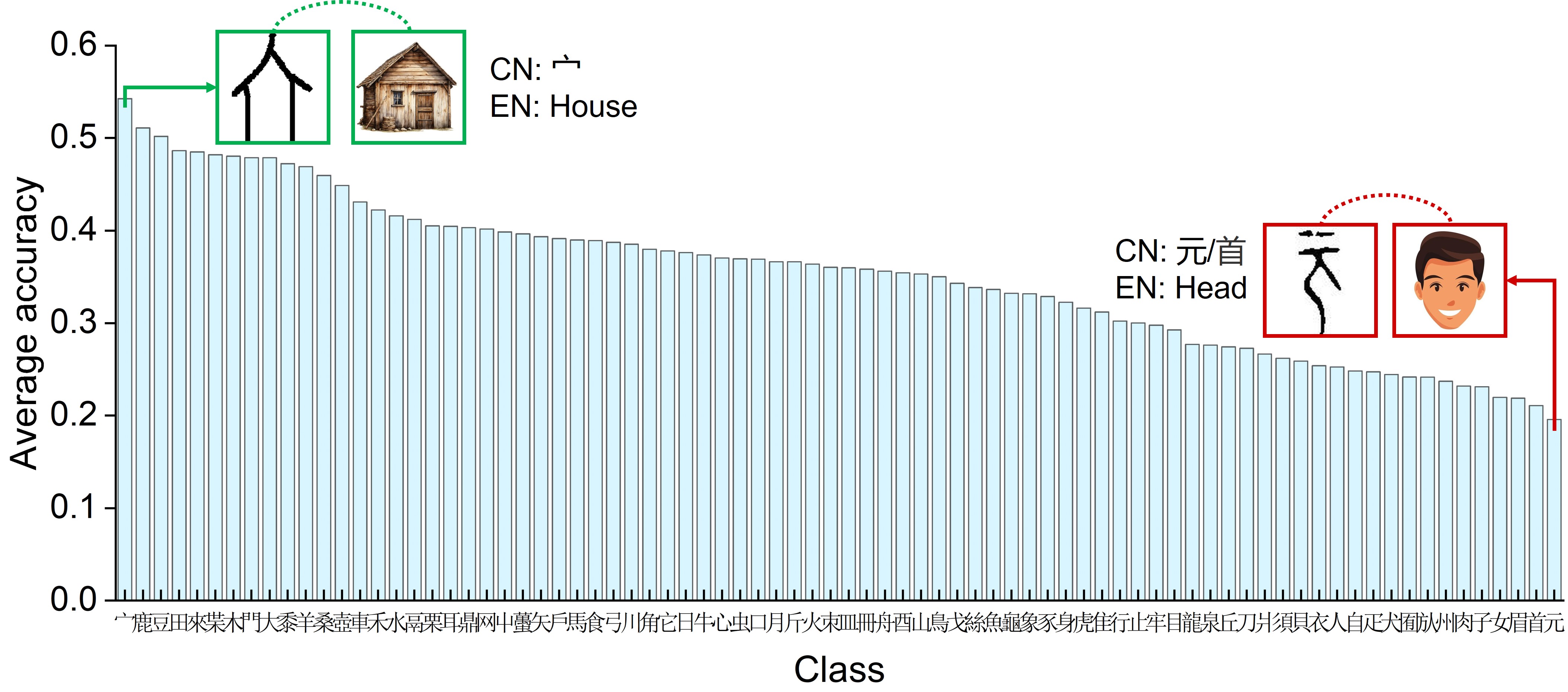}
\caption{Average accuracy in terms of OBC classes. We also illustrate the OBC classes with the highest and lowest accuracy.}
\label{acc_cate}
\end{center}
\end{figure}

\begin{figure}[t]
\begin{center}
\includegraphics[width=1\linewidth]{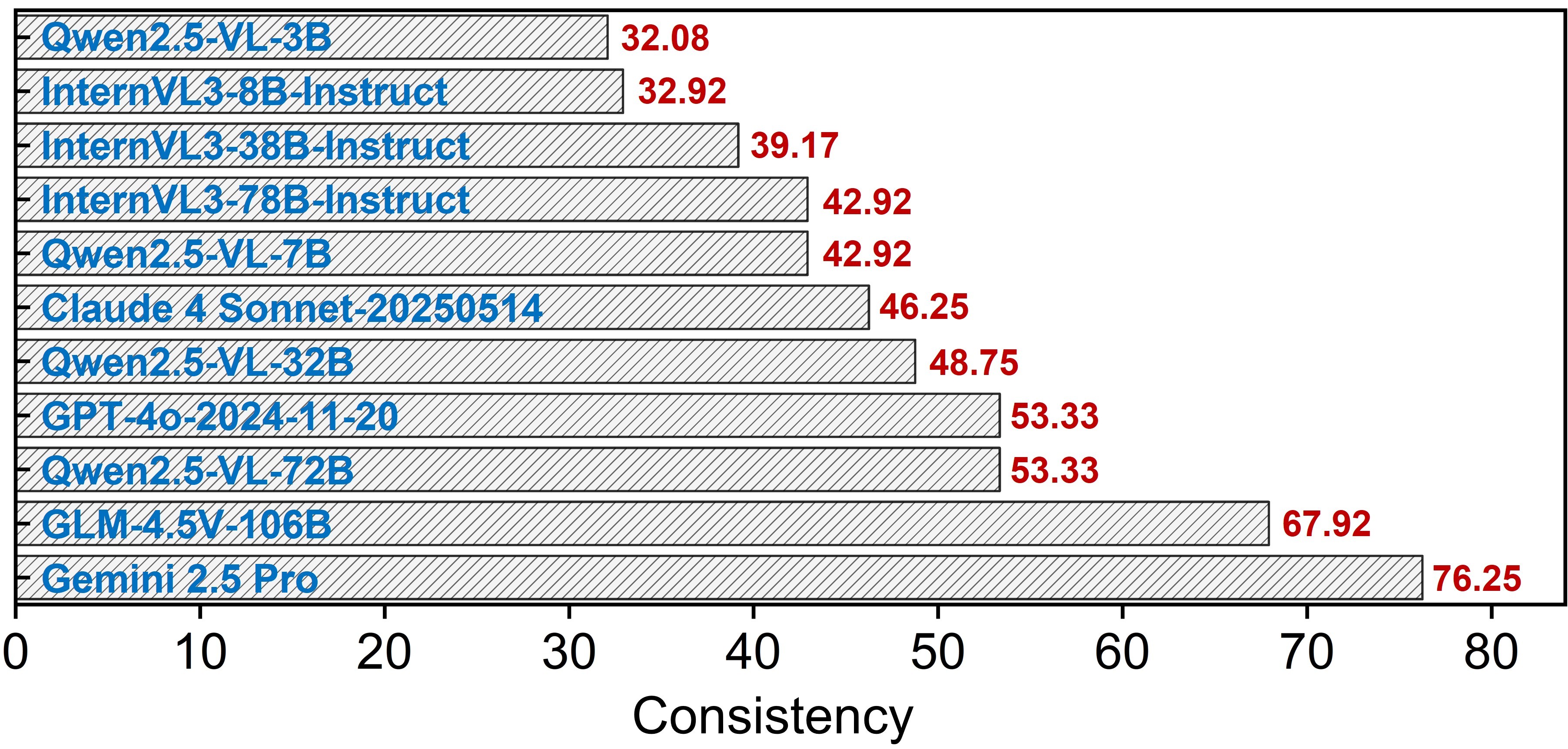}
\caption{The consistency (\%) of visual reference on 240 OBC-object pairs between humans and LMMs.}
\label{alignment}
\end{center}
\end{figure}

\begin{figure}[t]
\begin{center}
\includegraphics[width=1\linewidth]{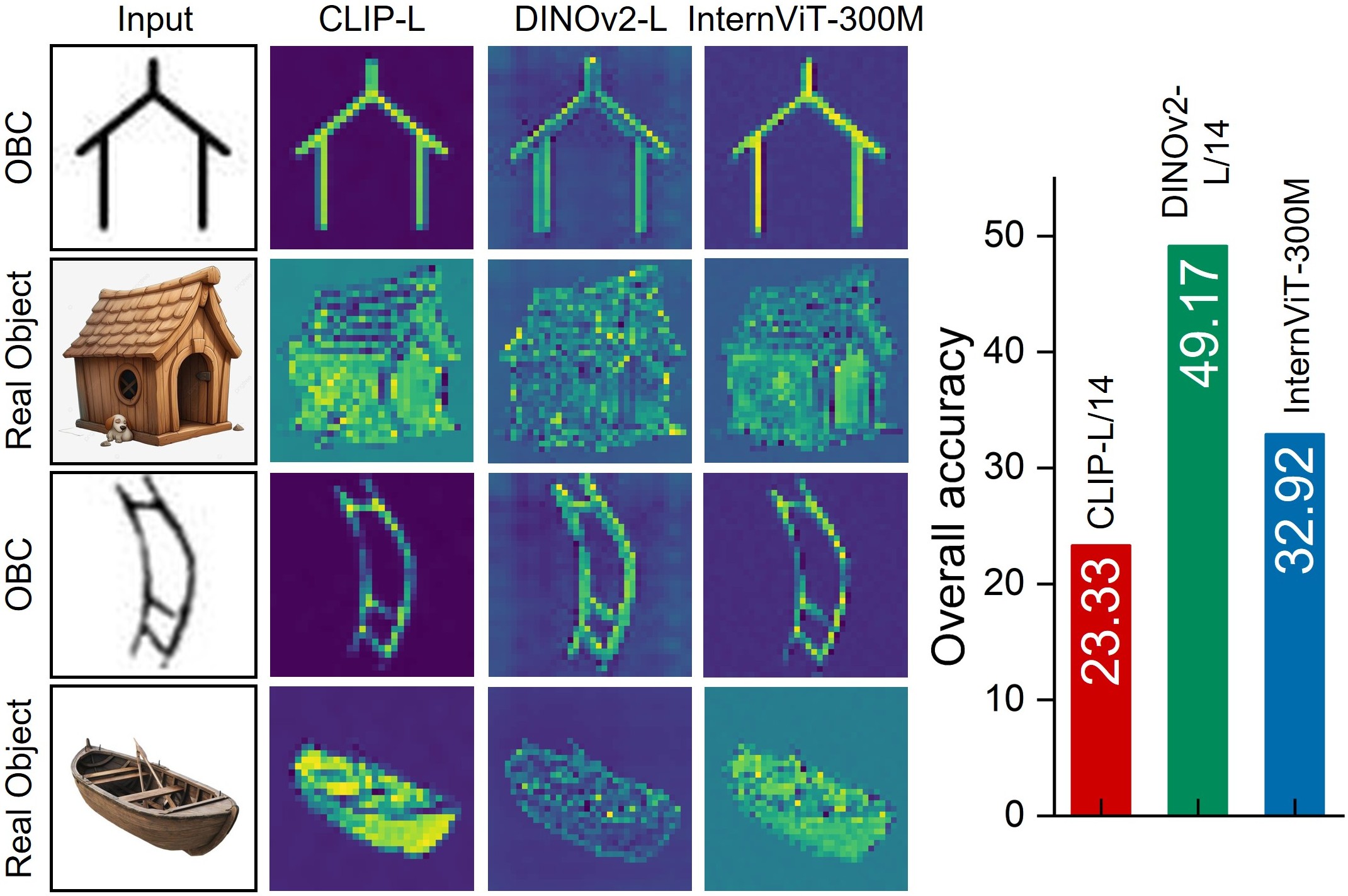}
\caption{Visualization of the attention maps of vision transformers ({\it left}) and accuracy supported by a direct readout of visual encoders ({\it right}).}
\label{attentionmap}
\end{center}
\end{figure}

\subsection{Main Results}

\noindent
{\bf Inferior performance of existing LMMs.} 
In Tab. \ref{performance}, we show the performance of 11 LMMs on the multi-choice question-based visual decipherment task. We observe that Gemini 2.5 Pro and InternVL3-38B-Instruct reach the top 2 accuracy and lead proprietary models and open-source models, respectively. Meanwhile, we notice that accuracies on complex OBCs are lower than on normal OBCs, indicating their discrepancy in the difficulty of visual decipherment. Moreover, we uncover a distinctive scaling law in both InternVL3 and Qwen2.5-VL series that exhibits a monotonic increase followed by saturation, i.e., an “increase-first, saturate-later” behavior. This suggests the existence of a potential “{\it golden}” ratio between the vision and language backbones, and reveals an inherent tension between visual representations and the language prior.
Overall, current LMMs still have difficulty in deciphering OBCs visually.

\noindent
{\bf Inter-class decipherment comparison.} 
In Fig. \ref{acc_cate}, we provide the accuracy distribution in terms of OBC classes. The average value and standard deviation of the accuracy for all OBC classes are 35.29\% and 8.25\%, respectively, ranging from 19.56\% to 54.25\%. We notice that the OBC class with the highest accuracy exhibits a high degree of structural similarity between the OBC and its corresponding object images, whereas such resemblance is rarely present in the class with the lowest accuracy, reflecting the nature of pictographic OBCs.

\noindent
{\bf Do LMMs behave human-like in visual decipherment?} We first compare the reference points of humans and LMMs in the visual alignment of OBC-object image pairs, shown in Fig. \ref{alignment}. Gemini 2.5 Pro shares the highest degree of consistency with humans (76.25\%). Besides, recent GLM-4.5V exceeds GPT-4o-20241120 significantly, showing better visual comparison ability. A typical scaling law is observed here.

\noindent
{\bf Effect of Vision Encoders.} We further disentangle LMMs and explore their vision backbones separately. As visualized in Fig. \ref{attentionmap}, the attention maps of DIVOv2 have more regional differences than CLIP-L and InternViT-300M. Some points with larger attention weights are more consistent with humans' reference points in Fig. \ref{annotated}. We also find the DIVOv2 performs even better than Claude 4 Sonnet and InternVL3-78B, suggesting that LMMs are not effectively using visual information or constrained by language priors.

\section{Conclusion}
In this work, we construct PictOBI-20k, a large-scale OBC-object paired image dataset for evaluating LMMs in terms of visual decipherment capabilities on pictographic OBCs. PictOBI-20k consists of 15k multi-choice questions and 240 reference point maps with accurate human annotations, which cover 80 OBC categories. We conduct a thorough evaluation of 11 prominent LMMs, comparing their performances and analyzing defects in their visual representations to provide insights for future research.



\end{document}